\documentclass[twoside,11pt]{article}

\usepackage{blindtext}

\usepackage{jmlr2e}
\usepackage[nolist]{acronym}
\usepackage{amsmath,graphicx}

\usepackage{algorithm}
\usepackage{algpseudocode}
\usepackage{amsmath,amssymb,amsfonts}
\usepackage{capt-of}
\usepackage{hyperref}
\usepackage{color}

\usepackage[capitalise]{cleveref}
\usepackage{longtable}
\usepackage{makecell}
\usepackage{mathtools}
\usepackage{multirow}
\usepackage{siunitx}
\DeclareSIUnit\pixel{px}
\usepackage{tabularx}
\usepackage{xspace}
\usepackage{verbatim}
\makeatletter
\DeclareRobustCommand\onedot{\futurelet\@let@token\@onedot}
\def\@onedot{\ifx\@let@token.\else.\null\fi\xspace}

\let\oldthefigure\thefigure
\let\oldthetable\thetable
\newcommand{\supplementaryfigures}{%
  \setcounter{figure}{0}
  \setcounter{table}{0}
  \renewcommand{\thefigure}{S\oldthefigure}
  \renewcommand{\thetable}{S\oldthetable}
}

\begin{acronym}
\acro{wsi}[WSI]{Whole Slide Image}
\acro{he}[H\&E]{Hematoxylin \& Eosin}
\acro{vlm}[VLM]{Vision-Language Model}
\acro{laion}[LAION]{Large-scale Artificial Intelligence Open Network}
\acro{hp}[HP]{Hyperplastic Polyp}
\acro{ssa}[SSA]{Sessile Serrated Adenoma}
\acro{peft}[PEFT]{Parameter Efficient Fine-Tuning}
\acro{dapt}[DAPT]{Domain-adaptive PreTraining}
\acro{tapt}[TAPT]{Task-adaptive PreTraining}
\acro{lora}[LoRA]{Low-Rank Adaptation}
\acro{coop}[CoOp]{Context Optimization}
\acro{vpt}[VPT]{Visial-Prompt Tuning}
\acro{dcis}[DCIS]{Ductal Carcinoma In Situ}
\acro{lcis}[LCIS]{Lobular Carcinoma In Situ}
\acro{csc}[CSC]{Class-Specific Context}
\end{acronym}

\newcommand{\citeblueref}[1]{\textcolor{blue}{(\cite{#1})}}

\usepackage{lastpage}
\jmlrheading{}{2025}{1-\pageref{LastPage}}{}{}{}{Jingna Qiu, Nishanth Jain, Jonas Ammeling, Marc Aubreville and Katharina Breininger}

\ShortHeadings{Effortless VLM Specialization in
Histopathology without Annotation}{Qiu et al.}
\firstpageno{1}

\begin{document}

\title{Effortless Vision-Language Model Specialization in
Histopathology without Annotation}

\author{\name Jingna Qiu \email jingna.qiu@fau.de 
       \AND
       \name Nishanth Jain \email nishanth.jain@fau.de \\
       \addr 
       Friedrich-Alexander-Universit\"{a}t Erlangen-N\"{u}rnberg, 
       Erlangen, Germany
       \AND
       \name Jonas Ammeling \email jonas.ammeling@thi.de\\
       \addr 
       Ingolstadt University of Applied Sciences. 
       Ingolstadt, Germany
       \AND
       \name Marc Aubreville \email marc.aubreville@hs-flensburg.de\\
       \addr 
       Flensburg University of Applied Sciences, 
       Flensburg, Germany
       \AND
       \name Katharina Breininger \email katharina.breininger@uni-wuerzburg.de\\
       \addr 
       Julius-Maximilians-Universit\"{a}t W\"{u}rzburg, 
       Würzburg, Germany}

\editor{}

\maketitle

\begin{abstract}
Recent advances in \acp{vlm} in histopathology, such as CONCH and QuiltNet, have demonstrated impressive zero-shot classification capabilities across various tasks. However, their general-purpose design may lead to suboptimal performance in specific downstream applications. While supervised fine-tuning methods address this issue, they require manually labeled samples for adaptation. This paper investigates annotation-free adaptation of \acp{vlm} through continued pretraining on domain- and task-relevant image-caption pairs extracted from existing databases. Our experiments on two \acp{vlm}, CONCH and QuiltNet, across three downstream tasks reveal that these pairs substantially enhance both zero-shot and few-shot performance. Notably, with larger training sizes, continued pretraining matches the performance of few-shot methods while eliminating manual labeling. Its effectiveness, task-agnostic design, and annotation-free workflow make it a promising pathway for adapting \acp{vlm} to new histopathology tasks. Code is available at \url{https://github.com/DeepMicroscopy/Annotation-free-VLM-specialization}.
\end{abstract}

\begin{keywords}
  vision-language models, histopathology, task adaptation
\end{keywords}

\acresetall
\section{Introduction}
\Acp{vlm} integrate images and textual descriptions to improve representation learning. Their success on natural image datasets has catalyzed their adaptation for histopathology image analysis. Several specialized \acp{vlm} tailored to histopathology have been developed, including PLIP~\citeblueref{huang2023visual}, QuiltNet~\citeblueref{ikezogwo2023quilt}, and CONCH~\citeblueref{lu2024visual}. PLIP and QuiltNet fine-tune CLIP~\citeblueref{radford2021learning}, with PLIP utilizing image-caption pairs sourced from Twitter discussions among pathologists, while QuiltNet leverages educational videos from YouTube and medical literature. On the other hand, CONCH builds on CoCa~\citeblueref{yu2022coca}, incorporating a captioning objective alongside CLIP's contrastive objectives and being trained on a dataset derived from PubMed articles and educational materials. Thanks to extensive and varied pretraining, these models exhibit robust zero-shot performance in histopathology image interpretation, including tile classification and downstream tasks like \ac{wsi} segmentation, achieved by aggregating tile-level results. 

\begin{figure}[t]
\centering  \centerline{\includegraphics[width=\textwidth]{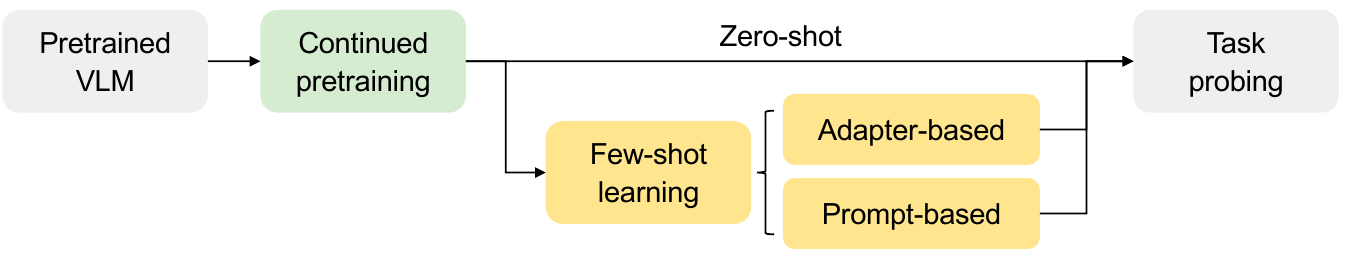}}
\caption{Retrieved image-caption pairs are utilized in the continued pretraining of a pretrained \ac{vlm} using a contrastive loss. Their effectiveness is evaluated in both zero-shot and few-shot learning scenarios, in comparison to the original \ac{vlm}.}
\label{fig:main}
\end{figure}

Nevertheless, training on broad and heterogeneous datasets may hinder a model's effectiveness on specialized tasks that necessitate more focused representations. To mitigate this issue, various supervised fine-tuning approaches have been proposed to refine model adaptation. Full fine-tuning updates all parameters of a pretrained encoder along with a classifier on top, while partial fine-tuning targets selective layers. \Ac{peft} methods introduce lightweight modules (e.g., adapters) that train only these components, keeping the pretrained weights fixed, thereby enhancing computational efficiency. Among these, LoRA~\citeblueref{hu2022lora} integrates trainable low-rank matrices in parallel with frozen dense layers to capture the residual for adjusting the original layer outputs. Adapter-based \ac{peft} strategies, such as CLIP-Adapter~\citeblueref{gao2024clip}, employ adapters that utilize the pre-trained image and text features and blend the adapter output as a residual to form the final features. Prompt-based adaptations, like \ac{coop}~\citeblueref{zhou2022learning}, involve training learnable prompt vectors instead of relying on hand-crafted prompts. 

However, all these methods necessitate manually annotated data for supervision, with \ac{peft} methods being relatively data-efficient and often employing few-shot datasets. Curating a labeled dataset requires specialized knowledge in pathology and becomes particularly challenging when identifying samples for rare diseases. This motivates our exploration of utilizing image-caption pairs from existing histopathology databases to enhance foundation model adaptation without the need for annotations. Specifically, we identify domain- and task-relevant samples through string matching. In the context of breast cancer classification, domain-relevant image-caption pairs are those where the caption includes the organ term ``breast'', while task-relevant pairs are a subset of the domain-specific pairs that contain one of the class names (e.g., ``normal'', ``benign'', ``in situ'', and ``invasive''). These domain- and task-specific pairs are subsequently used to continue the pretraining of the \ac{vlm} using a contrastive loss, referred to as \ac{dapt} and \ac{tapt}, respectively.

Our evaluation involves conducting \ac{dapt} and \ac{tapt} through full parameter updates and assessing the adapted model's performance on the targeted task in comparison to the original model. Furthermore, we investigate whether continued pretraining benefits subsequent few-shot learning methods, specifically \ac{coop}. Our experiments, encompassing three pathology tasks, including breast cancer classification on BACH~\citeblueref{aresta2019bach}, colorectal polyp classification on MHIST~\citeblueref{wei2021petri}, and prostate cancer grading on SICAP~\citeblueref{silva2020going}, demonstrate that the image-caption pairs identified from the retrieval dataset Quilt1M~\citeblueref{ikezogwo2023quilt} significantly enhance both zero-shot and few-shot performance of the foundation model in downstream tasks.


\section{Related Work}

\noindent\textbf{Continued Pretraining:} The methods of \ac{dapt} and \ac{tapt} have been evaluated using the language model RoBERTa~\citeblueref{liu2019roberta} in previous work~\citeblueref{gururangan2020don}, where relevant datasets were utilized as domain-specific data (e.g., Amazon reviews for the target task of IMDB review sentiment classification), while unlabeled training data (curated alongside the target task training set) served as task-specific data. In contrast, our work investigates the impact of retrieved image-caption pairs on \ac{vlm} adaptation, employing organ and class names as domain- and task-specific data, respectively.

\noindent\textbf{\ac{vlm} Adaptation via Fine-Tuning:} Prior research has predominantly focused on \ac{peft}, particularly using few-shot datasets. Among \textbf{adapter-based} methods, CLIP-Adapter~\citeblueref{gao2024clip} introduces an adapter consisting of two linear layers atop the text and image encoders, respectively, blending the adapter's output with the original embeddings to generate the final output. Tip-adapter~\citeblueref{zhang2022tip} is a cache-based adapter that calculates image embeddings for each few-shot training sample, integrating their one-hot labels weighted by their similarity to the input image during classification. Tip-X~\citeblueref{udandarao2023sus} further incorporates image-text similarities, given that the contrastive loss aims to align the two modalities. The cache modules in both Tip-adapter and Tip-X are fine-tuned using the few-shot training data. 
Among \textbf{prompt-based} methods, CoOp~\citeblueref{zhou2022learning} transforms context words in a prompt into a set of learnable vectors, replacing hand-crafted prompts such as ``A photo of \{class\}''. Meanwhile, CoCoOp~\citeblueref{zhou2022conditional} extends CoOp by implementing a lightweight module that generates an input-conditional token for each image. Additionally, CLIP-LoRA~\citeblueref{zanella2024low} applies \ac{lora} to CLIP by injecting low-rank matrices into the query, key, and value matrices within the attention block, maintaining a rank of $2$.


\section{Method}
Our goal is to evaluate the effectiveness of retrieved image-caption pairs in adapting pretrained histopathology \acp{vlm} to specific downstream tasks. These pairs are utilized to continue pretraining a foundational \ac{vlm} using a contrastive loss. We compare the performance of the adapted model to that of the original model in both zero-shot and few-shot learning scenarios. The workflow is illustrated in \cref{fig:main}.

\subsection{Image-Caption Pair Retrieval}
To conduct \ac{dapt} and \ac{tapt}, we collect domain- and task-specific image-caption pairs via string matching from existing histopathology image-caption databases. The aim is to determine whether adapting the model by exposing it to general domain information (e.g., regarding organs) or more specialized information on targeted classes is more beneficial. Domain-specific image-caption pairs are identified if the caption includes one of the site keywords, such as the name of the organ. In specific cases where a certain subject is involved in the task, such as in colorectal polyp classification, the subject ``polyp'' also serves as a keyword. Task-specific pairs are filtered as a subset of the domain-specific pairs using class names pertinent to the task. Synonyms and alternative terms, such as \ac{dcis} and \ac{lcis} for the class "in situ" in breast cancer classification, are excluded to avoid reliance on specialized medical knowledge. Full names are employed when class names involve abbreviations.

Once identified, the pairs are ranked based on the alignment of image-caption pairs using the CONCH model due to its general high performance in zero-shot classification among histopathology tasks. Alignment scores are computed as the cosine similarity between normalized image embeddings \(x_i\) and caption embeddings \(y_i\) for the \(i^{\text{th}}\) pair, as
\begin{equation}
    sim(x_i, y_i) = x_i^\top y_i.
\end{equation}
This ranking prioritizes high-quality pairs when using a limited training set and facilitates filtering out noisy data by discarding poorly related pairs.

We opted for a straightforward string matching method, as preliminary experiments showed that it leads to higher-quality retrievals than similarity searches based on embeddings from PathologyBERT~\citeblueref{santos2023pathologybert} or caption classification using Gemma3~\citeblueref{kamath2025gemma}.

\subsection{Continued Pretraining}
We proceed with continued pretraining using the domain-specific image-caption pairs in \ac{dapt} and the task-specific pairs in \ac{tapt}. Both the image and text encoders in the \ac{vlm} are updated using a dual-encoder contrastive loss~\citeblueref{yu2022coca}, as
\begin{equation}
    L_{\text{contrast}} = -\frac{1}{N} \left( \sum_{i=1}^{N} \log \frac{\exp(x_i^\top y_i)}{\sum_{j=1}^{N} \exp(x_i^\top y_j)} + \sum_{i=1}^{N} \log \frac{\exp(y_i^\top x_i)}{\sum_{j=1}^{N} \exp(y_i^\top x_j)} \right),
\end{equation}
where \(N\) is the batch size. The cosine similarities of the paired embeddings are maximized relative to other negative pairings within the batch.

\subsection{Evaluation of Continued Pretraining Effect}
To assess the impact of continued pretraining, we evaluate the model in both zero-shot and few-shot scenarios. In zero-shot classification, the text encoder computes embeddings for each class text prompt, and the class with the highest cosine similarity is assigned to the test image. We utilize the same multiple class descriptions and prompt templates employed in CONCH~\citeblueref{lu2024visual} for prompt ensembling, ensuring fair comparisons. 

For few-shot learning, we adopt \ac{coop} due to its established effectiveness. In \ac{coop}, prompt vectors are learned to substitute hand-crafted prompts for the text encoder input, while the pretrained text encoder remains frozen. We conduct experiments with two types of prompts: a unified context prompt \( [V]_1[V]_2[V]_3 \ldots [V]_M[CLASS] \) and a \ac{csc} prompt \( [V]^c_1[V]^c_2[V]^c_3 \ldots [V]^c_M \) unique to each class \(c\). Here, \( [V]_M \) denotes a context token of the same dimension as the word embeddings and $M$ is the context length. It has been shown that unified and \ac{csc} prompts yield better performance in generic and fine-grained object classification on natural images, respectively, and shorter context lengths enhance generalization while longer ones improve performance~\citeblueref{zhou2022learning}. Following the original paper, we experiment with both context types and lengths of $4$ and $16$ to determine the optimal combination for histopathology tasks.

\section{Experimental Setups}

\subsection{Image-caption Source Database Quilt1M}
The Quilt1M database~\citeblueref{ikezogwo2023quilt} contains $1,017,708$ image-caption pairs collected from multiple sources: $802,144$ pairs from $1,087$ hours of educational histopathology YouTube videos, $59,371$ pairs from PubMed open-access articles, $22,682$ histopathology-related pairs from the LAION-5B dataset~\citeblueref{schuhmann2022laion}, and $133,511$ pairs from $55,000$ curated tweets in OpenPath~\citeblueref{huang2023visual}. Extracting image-caption pairs from YouTube videos presents challenges such as accurate speech-to-text conversion, particularly for medical terminology, frame extraction and precise alignment of image frames with corresponding text. Due to the dataset's scale, manual verification is impractical. Aubreville et al.~\citeblueref{aubreville2024model} refined Quilt1M by removing low-quality images and those containing extraneous elements such as narrators or overlaid text. This process resulted in a cleaned subset of $232,039$ image-caption pairs, which we use in this study.

\subsection{Histopathology-specific Vision-language Models}
\noindent\textbf{CONCH} is a CoCa-based \ac{vlm} trained on an in-house dataset of over $1.17$ million image-caption pairs sourced from PubMed and educational materials. \noindent\textbf{QuiltNet} is a CLIP-based \ac{vlm} trained on the Quilt1M database. The selection of these two models aims to understand the impact of whether the retrieval dataset Quilt1M was included in the original foundation model's pretraining process.


\begin{table}[t]
\centering
\caption{Keywords for domain- and task-specific image-caption pairs retrieval and the retrieved pair amounts for \ac{dapt} and \ac{tapt}.}
\begin{tabularx}{\linewidth}{|m{1.5cm}|m{3cm}|X|m{1cm}|m{1cm}|}
\hline
Task & Site keywords & Class keywords & DAPT & TAPT \tabularnewline \hline
BACH & breast & normal, benign, in situ, invasive & 5437 & 896 \tabularnewline \hline
MHIST & colon, colorectal, polyp & hyperplastic, benign, sessile, serrated, adenoma & 5224 & 806 \tabularnewline \hline
SICAP & prostate, gland & non-cancerous, Gleason & 10749 & 154 \tabularnewline \hline
\end{tabularx}
\label{tab:keywords}
\end{table}

\subsection{Downstream Tasks and Datasets}
\noindent\textbf{Breast Cancer Classification on BACH} The BACH dataset~\citeblueref{aresta2019bach} contains microscopy images used for a breast cancer subtyping task across four classes: normal, benign, in situ carcinoma, and invasive carcinoma. The dataset comprises $400$ images, with 100 images per class. Each image has a standardized resolution of $2048\times1536$ \si{pixels} at $0.42$~\si{\micro\meter\per\pixel}.

\noindent\textbf{Colorectal Polyps Classification on MHIST}
The MHIST dataset~\citeblueref{wei2021petri} comprises $3152$ images ($2162$ \acp{hp}, $990$ \acp{ssa}) extracted from $328$ \acp{wsi}, each with a resolution of $224\times224$ \si{pixels} at $8\times$ magnification. The test set of $977$ images are used for evaluation.  

\noindent\textbf{Prostate Cancer Grading on SICAP}
The SICAP dataset~\citeblueref{silva2020going} is used for prostate cancer diagnosis based on the Gleason grading system~\citeblueref{gleason1992histologic}. 
SICAP contains $10340$ image patches extracted from $182$ \acp{wsi} ($4417$ non-cancerous, $1636$ grade $3$, $3622$ grade $4$, $665$ grade $5$). Each patch has a resolution of $512\times512$ \si{pixels} at $10\times$ magnification. The test set of $2122$ images are used for evaluation.

A complete list of keywords used to identify relevant domain- and task-specific image-caption pairs for all datasets is provided in~\cref{tab:keywords}.

\subsection{Evaluation Metrics and Implementation Details}
Balanced accuracy is used to evaluate the experiments on BACH and MHIST, while Cohen’s quadratic kappa is used for SICAP, following~\citeblueref{lu2024visual}. 

To assess the impact of training size $N$ on continued pretraining, we select the first $N$ pairs in the sorted retrievals (according to their CONCH alignment score). We adopt the convention of using ``shots'' to denote the training size, defined as $N=shots \times num\_classes$, following few-shot learning conventions. Continued pretraining experiments perform full parameter update and use contrastive loss. Few-shot experiments perform \ac{coop} and use cross-entropy loss. All experiments use the AdamW optimizer with cosine annealing as the learning rate scheduler for $50$ epochs, with initial learning rate and weight decay parameters being tuned for each training size based on the minimal training loss after $5$ epochs.
\begin{figure}[t]
  \centering  \centerline{\includegraphics[width=\textwidth]{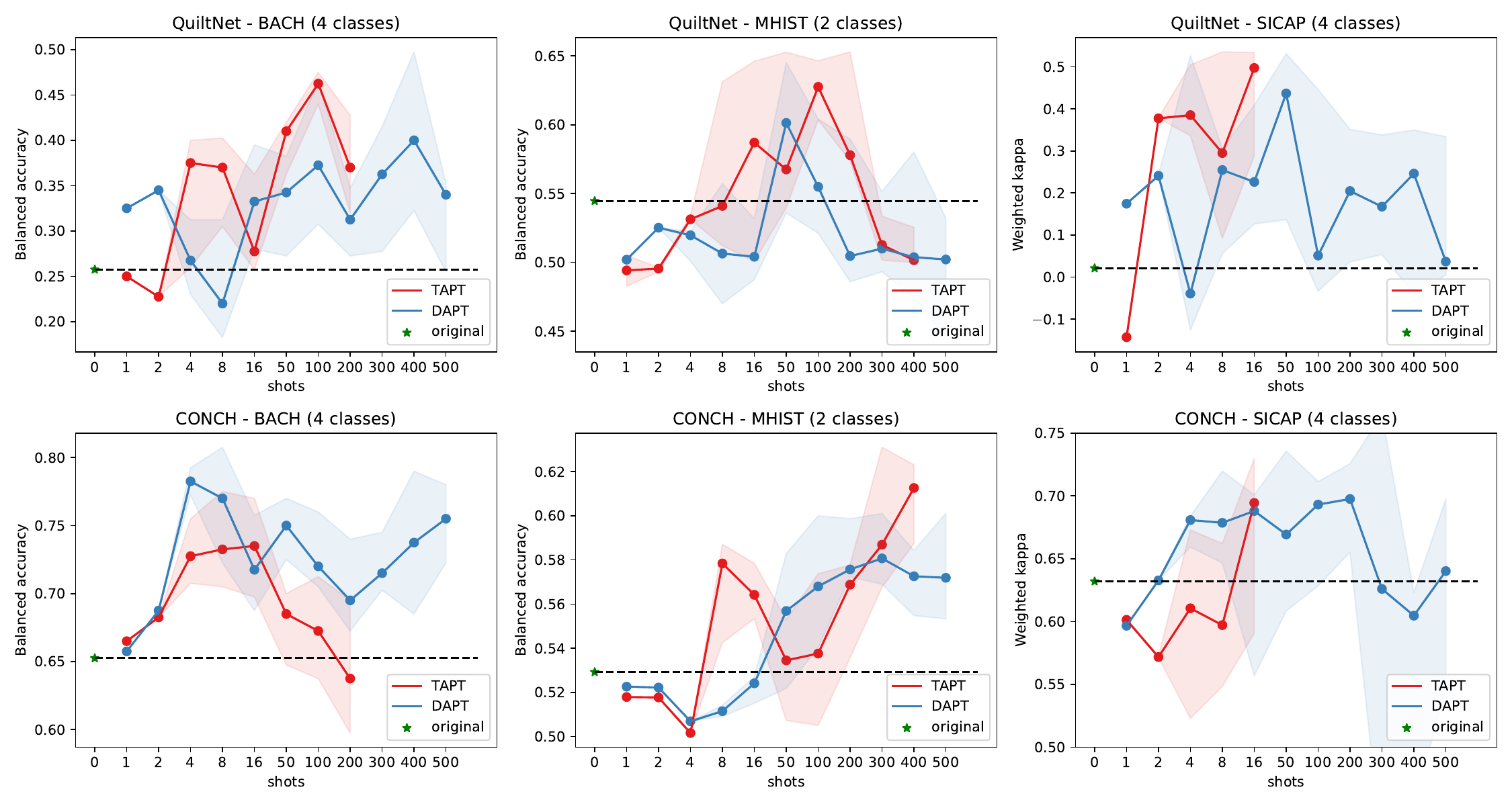}}
\caption{Comparison of continued pretraining methods, \ac{dapt} and \ac{tapt}, with the original unadapted model on QuiltNet (top row) and CONCH (bottom row) across three pathology tasks up to $500$ shots. The performance of \ac{tapt} may terminate earlier if insufficient pairs are retrieved. The results show the median values from five repetitions, with minimum and maximum values shaded.}
\label{fig:dapt_tapt}
\end{figure}

\begin{figure}[t]
  \centering  \centerline{\includegraphics[width=\textwidth]{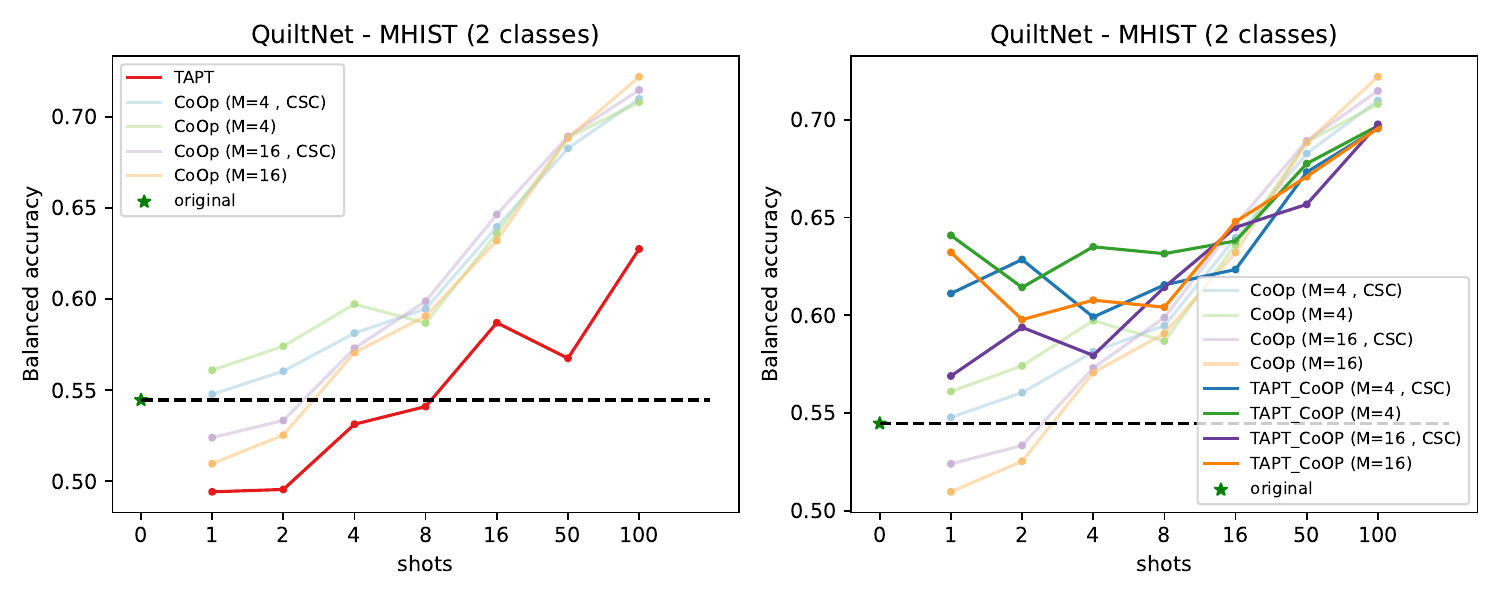}}
\caption{
Left: Comparison of \ac{tapt} and \ac{coop} using unlabeled and labeled training data, respectively. $M$ denotes the prompt's context length. \ac{csc} indicates the use of class-specific context, while the unified context is used otherwise. Right: Comparison of \ac{coop} with and without \ac{tapt} across various context types and lengths. Results present the median values from ten randomly selected few-shot sets.}
\label{fig:coop}
\end{figure}
\section{Results and Discussion}
\subsection{Effect of Continued Pretraining on Zero-shot}
\Cref{fig:dapt_tapt} compares the performance of \ac{dapt} and \ac{tapt} across various training sizes. \Ac{dapt} (blue) demonstrates consistent improvements in zero-shot performance across all tasks and models compared to the original unadapted model (black dash line), except for MHIST when applied to QuiltNet. The benefit of increasing the number of training image-caption pairs on \ac{dapt} is not pronounced, except in the case of MHIST when applied to CONCH. Moreover, performance may decline after a certain amount of training data is used, as observed on SICAP. Given that the training pairs are sorted by alignment score according to CONCH, it is likely that lower-ranked pairs may detrimentally affect model performance. In contrast, \ac{tapt} (red) achieves more stable improvements as additional training pairs are utilized. When applied to QuiltNet, \ac{tapt} surpasses \ac{dapt}, enhancing the original model's performance by $79.62\%$ and $15.22\%$ on BACH and MHIST, respectively, compared to the $55.34\%$ and $10.41\%$ improvements achieved by \ac{dapt}. Furthermore, \ac{tapt} significantly increases the weighted Cohen's kappa on SICAP from $0.02$ to $0.50$, achieving higher performance improvement than \ac{dapt} with fewer training pairs. Interestingly, when moving to CONCH, \ac{dapt} exhibits higher and more consistent performance improvements across datasets, especially on BACH and SICAP, suggesting that domain adaptation via large-scale caption-text alignment benefits CONCH more strongly. This difference may arise because QuiltNet has already been pre-trained on Quilt1M~\citeblueref{ikezogwo2023quilt}, making domain information in the training pairs redundant, whereas CONCH, without prior exposure to Quilt1M, benefits more from \ac{dapt}. Overall, the use of retrieved domain- or task-specific image-text pairs demonstrates promising potential for model adaptation, yielding more substantial improvements particularly when the baseline performance of the original \ac{vlm} is low. This is especially noteworthy given that the retrieval database, Quilt1M, is primarily sourced from uncurated YouTube videos and lacks manual verification, which underscores the robustness of the adaptation process of continued pretraining even in the presence of noisy retrieval sources. Additional results utilizing \ac{lora} for continued pretraining as a replacement for full parameter updates, along with the use of PathologyBERT and Gemma3 models for assigning pseudo-labels to the retrieved pairs to balance the training set, are presented in the supplementary material.

\subsection{Effect of Continued Pretraining on Few-shot}
Due to the relatively poorer performance improvements of \ac{dapt} and \ac{tapt} on MHIST with QuiltNet in zero-shot evaluations compared to other datasets and models, we apply the few-shot method \ac{coop} to assess whether the adapted model's performance aligns with that of few-shot learning. Additionally, we investigate whether applying \ac{coop} on top of the adapted model yields higher performance than the original unadapted model. We repeat with $10$ randomly selected few-shot sets from the MHIST training set and report the median values from the evaluation on the test set (see Fig.~\ref{fig:coop}). Our observations indicate that a context length of $4$ with a unified context prompt performs the best up to $4$ shots for \ac{coop} on MHIST, showing minimal differences thereafter with varying context lengths and types. Notably, the performance improvements to the original \ac{vlm} achieved through \ac{coop} can also be attained with \ac{tapt} when a larger training size with no need for manual annotation. For instance, \ac{tapt} achieves performance comparable to \ac{coop} at $8$ shots by utilizing $16$ shots, and a performance similar to \ac{coop} at $16$ shots with $100$ shots (Fig.~\ref{fig:coop} left). Further, with up to 8 shots, \ac{tapt} significantly enhances \ac{coop} by achieving higher performance compared to its application on the original unadapted model (Fig.~\ref{fig:coop} right).

\section{Conclusion and Future Work}
We demonstrated that continued pretraining of a \ac{vlm} with retrieved domain- and task-specific image–caption pairs enhances its performance in both zero-shot and few-shot learning settings. Our method achieves results comparable to the few-shot learning approach \ac{coop} when trained on larger datasets, while eliminating the need for manual annotation. Its label-free and task-agnostic design offers a promising direction for adapting foundation \acp{vlm} to new tasks.

Future work includes optimizing batch composition strategies, such as avoiding batches containing images with highly similar texture descriptions, to stabilize contrastive loss convergence. Applying a threshold on CONCH alignment scores could filter out poorly aligned pairs and prevent performance degradation. While CONCH scores effectively proxy pair quality, ensembling alignment measurements from multiple \acp{vlm} may further enhance data selection. Integrating our annotation-free adaptation with few-shot learning by ranking pairs based on similarity to few-shot images could improve training relevance and facilitate more meaningful clinical deployment.


\acks{We acknowledge support by d.hip campus - Bavarian aim (J.Q.), the German Research Foundation (DFG) project 460333672 CRC1540 EBM, project 405969122 FOR2886 Pandora, as well as the scientific support and HPC resources provided by the Erlangen National High Performance Computing Center (NHR@FAU) of the Friedrich-Alexander-Universität Erlangen-Nürnberg (FAU). NHR funding is provided by federal and Bavarian state authorities. NHR@FAU hardware is partially funded by the German Research Foundation (DFG) – 440719683.}


\vskip 0.2in
\bibliography{main}

\begin{thebibliography}{22}
\providecommand{\natexlab}[1]{#1}
\providecommand{\url}[1]{\texttt{#1}}
\expandafter\ifx\csname urlstyle\endcsname\relax
  \providecommand{\doi}[1]{doi: #1}\else
  \providecommand{\doi}{doi: \begingroup \urlstyle{rm}\Url}\fi

\bibitem[Aresta et~al.(2019)Aresta, Ara{\'u}jo, Kwok, Chennamsetty, Safwan, Alex, Marami, Prastawa, Chan, Donovan, et~al.]{aresta2019bach}
Guilherme Aresta, Teresa Ara{\'u}jo, Scotty Kwok, Sai~Saketh Chennamsetty, Mohammed Safwan, Varghese Alex, Bahram Marami, Marcel Prastawa, Monica Chan, Michael Donovan, et~al.
\newblock Bach: Grand challenge on breast cancer histology images.
\newblock \emph{Medical image analysis}, 56:\penalty0 122--139, 2019.

\bibitem[Aubreville et~al.(2024)Aubreville, Ganz, Ammeling, Kaltenecker, and Bertram]{aubreville2024model}
Marc Aubreville, Jonathan Ganz, Jonas Ammeling, Christopher~C Kaltenecker, and Christof~A Bertram.
\newblock Model-based cleaning of the quilt-1m pathology dataset for text-conditional image synthesis.
\newblock \emph{Medical Imaging with Deep Learning (MIDL)}, 2024.
\newblock URL \url{https://openreview.net/forum?id=m7wYKrUjzV}.

\bibitem[Gao et~al.(2024)Gao, Geng, Zhang, Ma, Fang, Zhang, Li, and Qiao]{gao2024clip}
Peng Gao, Shijie Geng, Renrui Zhang, Teli Ma, Rongyao Fang, Yongfeng Zhang, Hongsheng Li, and Yu~Qiao.
\newblock Clip-adapter: Better vision-language models with feature adapters.
\newblock \emph{International Journal of Computer Vision}, 132\penalty0 (2):\penalty0 581--595, 2024.

\bibitem[Gleason(1992)]{gleason1992histologic}
Donald~F Gleason.
\newblock Histologic grading of prostate cancer: a perspective.
\newblock \emph{Human pathology}, 23\penalty0 (3):\penalty0 273--279, 1992.

\bibitem[Gururangan et~al.(2020)Gururangan, Marasovi{\'c}, Swayamdipta, Lo, Beltagy, Downey, and Smith]{gururangan2020don}
Suchin Gururangan, Ana Marasovi{\'c}, Swabha Swayamdipta, Kyle Lo, Iz~Beltagy, Doug Downey, and Noah~A Smith.
\newblock Don't stop pretraining: Adapt language models to domains and tasks.
\newblock \emph{arXiv preprint arXiv:2004.10964}, 2020.

\bibitem[Hu et~al.(2022)Hu, Shen, Wallis, Allen-Zhu, Li, Wang, Wang, Chen, et~al.]{hu2022lora}
Edward~J Hu, Yelong Shen, Phillip Wallis, Zeyuan Allen-Zhu, Yuanzhi Li, Shean Wang, Lu~Wang, Weizhu Chen, et~al.
\newblock Lora: Low-rank adaptation of large language models.
\newblock \emph{ICLR}, 1\penalty0 (2):\penalty0 3, 2022.

\bibitem[Huang et~al.(2023)Huang, Bianchi, Yuksekgonul, Montine, and Zou]{huang2023visual}
Zhi Huang, Federico Bianchi, Mert Yuksekgonul, Thomas~J Montine, and James Zou.
\newblock A visual--language foundation model for pathology image analysis using medical twitter.
\newblock \emph{Nature medicine}, 29\penalty0 (9):\penalty0 2307--2316, 2023.

\bibitem[Ikezogwo et~al.(2023)Ikezogwo, Seyfioglu, Ghezloo, Geva, Sheikh~Mohammed, Anand, Krishna, and Shapiro]{ikezogwo2023quilt}
Wisdom Ikezogwo, Saygin Seyfioglu, Fatemeh Ghezloo, Dylan Geva, Fatwir Sheikh~Mohammed, Pavan~Kumar Anand, Ranjay Krishna, and Linda Shapiro.
\newblock Quilt-1m: One million image-text pairs for histopathology.
\newblock \emph{Advances in neural information processing systems}, 36:\penalty0 37995--38017, 2023.

\bibitem[Kamath et~al.(2025)Kamath, Ferret, Pathak, Vieillard, Merhej, Perrin, Matejovicova, Ram{\'e}, Rivi{\`e}re, Rouillard, et~al.]{kamath2025gemma}
Aishwarya Kamath, Johan Ferret, Shreya Pathak, Nino Vieillard, Ramona Merhej, Sarah Perrin, Tatiana Matejovicova, Alexandre Ram{\'e}, Morgane Rivi{\`e}re, Louis Rouillard, et~al.
\newblock Gemma 3 technical report.
\newblock \emph{CoRR}, 2025.

\bibitem[Liu et~al.(2019)Liu, Ott, Goyal, Du, Joshi, Chen, Levy, Lewis, Zettlemoyer, and Stoyanov]{liu2019roberta}
Yinhan Liu, Myle Ott, Naman Goyal, Jingfei Du, Mandar Joshi, Danqi Chen, Omer Levy, Mike Lewis, Luke Zettlemoyer, and Veselin Stoyanov.
\newblock Roberta: A robustly optimized bert pretraining approach.
\newblock \emph{arXiv preprint arXiv:1907.11692}, 2019.

\bibitem[Lu et~al.(2024)Lu, Chen, Williamson, Chen, Liang, Ding, Jaume, Odintsov, Le, Gerber, et~al.]{lu2024visual}
Ming~Y Lu, Bowen Chen, Drew~FK Williamson, Richard~J Chen, Ivy Liang, Tong Ding, Guillaume Jaume, Igor Odintsov, Long~Phi Le, Georg Gerber, et~al.
\newblock A visual-language foundation model for computational pathology.
\newblock \emph{Nature Medicine}, 30\penalty0 (3):\penalty0 863--874, 2024.

\bibitem[Radford et~al.(2021)Radford, Kim, Hallacy, Ramesh, Goh, Agarwal, Sastry, Askell, Mishkin, Clark, et~al.]{radford2021learning}
Alec Radford, Jong~Wook Kim, Chris Hallacy, Aditya Ramesh, Gabriel Goh, Sandhini Agarwal, Girish Sastry, Amanda Askell, Pamela Mishkin, Jack Clark, et~al.
\newblock Learning transferable visual models from natural language supervision.
\newblock In \emph{International conference on machine learning}, pages 8748--8763. PmLR, 2021.

\bibitem[Santos et~al.(2023)Santos, Tariq, Das, Vayalpati, Smith, Trivedi, and Banerjee]{santos2023pathologybert}
Thiago Santos, Amara Tariq, Susmita Das, Kavyasree Vayalpati, Geoffrey~H Smith, Hari Trivedi, and Imon Banerjee.
\newblock Pathologybert-pre-trained vs. a new transformer language model for pathology domain.
\newblock In \emph{AMIA annual symposium proceedings}, volume 2022, page 962, 2023.

\bibitem[Schuhmann et~al.(2022)Schuhmann, Beaumont, Vencu, Gordon, Wightman, Cherti, Coombes, Katta, Mullis, Wortsman, et~al.]{schuhmann2022laion}
Christoph Schuhmann, Romain Beaumont, Richard Vencu, Cade Gordon, Ross Wightman, Mehdi Cherti, Theo Coombes, Aarush Katta, Clayton Mullis, Mitchell Wortsman, et~al.
\newblock Laion-5b: An open large-scale dataset for training next generation image-text models.
\newblock \emph{Advances in neural information processing systems}, 35:\penalty0 25278--25294, 2022.

\bibitem[Silva-Rodr{\'\i}guez et~al.(2020)Silva-Rodr{\'\i}guez, Colomer, Sales, Molina, and Naranjo]{silva2020going}
Julio Silva-Rodr{\'\i}guez, Adri{\'a}n Colomer, Mar{\'\i}a~A Sales, Rafael Molina, and Valery Naranjo.
\newblock Going deeper through the gleason scoring scale: An automatic end-to-end system for histology prostate grading and cribriform pattern detection.
\newblock \emph{Computer methods and programs in biomedicine}, 195:\penalty0 105637, 2020.

\bibitem[Udandarao et~al.(2023)Udandarao, Gupta, and Albanie]{udandarao2023sus}
Vishaal Udandarao, Ankush Gupta, and Samuel Albanie.
\newblock Sus-x: Training-free name-only transfer of vision-language models.
\newblock In \emph{Proceedings of the IEEE/CVF International Conference on Computer Vision}, pages 2725--2736, 2023.

\bibitem[Wei et~al.(2021)Wei, Suriawinata, Ren, Liu, Lisovsky, Vaickus, Brown, Baker, Tomita, Torresani, et~al.]{wei2021petri}
Jerry Wei, Arief Suriawinata, Bing Ren, Xiaoying Liu, Mikhail Lisovsky, Louis Vaickus, Charles Brown, Michael Baker, Naofumi Tomita, Lorenzo Torresani, et~al.
\newblock A petri dish for histopathology image analysis.
\newblock In \emph{Artificial Intelligence in Medicine: 19th International Conference on Artificial Intelligence in Medicine, AIME 2021, Virtual Event, June 15--18, 2021, Proceedings}, pages 11--24. Springer, 2021.

\bibitem[Yu et~al.(2022)Yu, Wang, Vasudevan, Yeung, Seyedhosseini, and Wu]{yu2022coca}
Jiahui Yu, Zirui Wang, Vijay Vasudevan, Legg Yeung, Mojtaba Seyedhosseini, and Yonghui Wu.
\newblock Coca: Contrastive captioners are image-text foundation models.
\newblock \emph{arXiv preprint arXiv:2205.01917}, 2022.

\bibitem[Zanella and Ben~Ayed(2024)]{zanella2024low}
Maxime Zanella and Ismail Ben~Ayed.
\newblock Low-rank few-shot adaptation of vision-language models.
\newblock In \emph{Proceedings of the IEEE/CVF Conference on Computer Vision and Pattern Recognition}, pages 1593--1603, 2024.

\bibitem[Zhang et~al.(2022)Zhang, Zhang, Fang, Gao, Li, Dai, Qiao, and Li]{zhang2022tip}
Renrui Zhang, Wei Zhang, Rongyao Fang, Peng Gao, Kunchang Li, Jifeng Dai, Yu~Qiao, and Hongsheng Li.
\newblock Tip-adapter: Training-free adaption of clip for few-shot classification.
\newblock In \emph{European conference on computer vision}, pages 493--510. Springer, 2022.

\bibitem[Zhou et~al.(2022{\natexlab{a}})Zhou, Yang, Loy, and Liu]{zhou2022conditional}
Kaiyang Zhou, Jingkang Yang, Chen~Change Loy, and Ziwei Liu.
\newblock Conditional prompt learning for vision-language models.
\newblock In \emph{Proceedings of the IEEE/CVF conference on computer vision and pattern recognition}, pages 16816--16825, 2022{\natexlab{a}}.

\bibitem[Zhou et~al.(2022{\natexlab{b}})Zhou, Yang, Loy, and Liu]{zhou2022learning}
Kaiyang Zhou, Jingkang Yang, Chen~Change Loy, and Ziwei Liu.
\newblock Learning to prompt for vision-language models.
\newblock \emph{International Journal of Computer Vision}, 130\penalty0 (9):\penalty0 2337--2348, 2022{\natexlab{b}}.

\end{thebibliography}

\newpage
\setcounter{page}{1}
\setcounter{section}{0}
\section{Supplementary Material}
\begin{figure}
\begin{minipage}[b]{\linewidth}
  \centering  \centerline{\includegraphics[width=\textwidth]{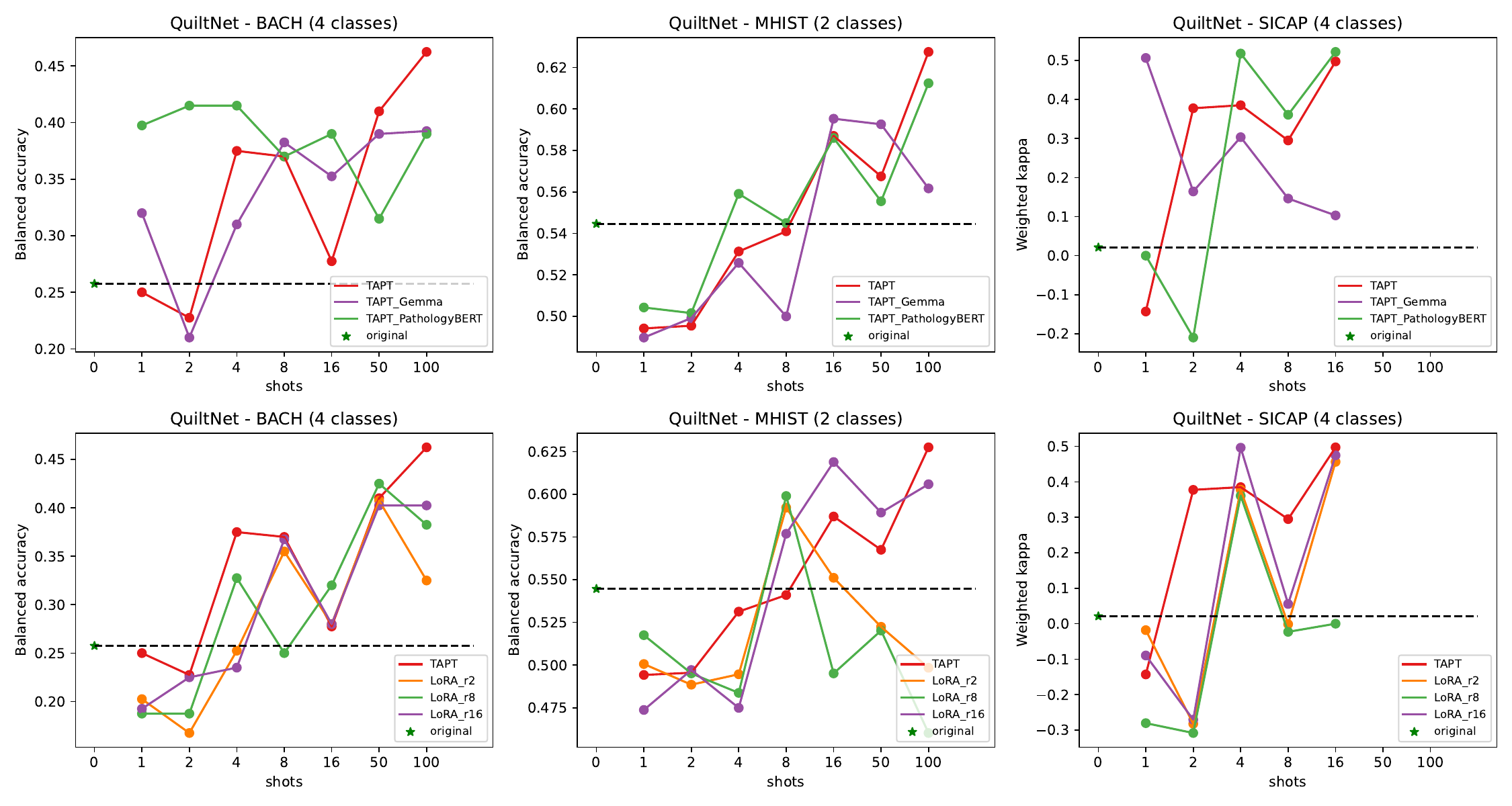}}
\end{minipage}
\caption{Top row: Effect of leveraging PathologyBERT and Gemma3 for balancing the training set. Bottom row: Effect of replacing full parameter update with \ac{lora}.}
\label{fig:TAPT_balanced_LoRA}
\end{figure}

\supplementaryfigures
We investigate the effect of balancing the training set among classes on QuiltNet by leveraging PathologyBERT and Gemma3 models for assigning pseudo-labels. As shown in~\cref{fig:TAPT_balanced_LoRA} (top row), we do not observe significant improvements from balancing, nor do we see better performance with Gemma3 classifications, which have a higher correctness rate compared to the labels provided by PathologyBERT. Additionally, we utilize \ac{lora} for continued pretraining as a replacement for full parameter updates to evaluate the generalization of the retrieved image-caption pairs. As shown in~\cref{fig:TAPT_balanced_LoRA} (bottom row), we found similar patterns of performance improvement compared to full parameter update. However, a lower rank (e.g., 2) can lead to decreased performance compared to full parameter update when more training data is employed. 
\end{document}